\documentclass[letterpaper, 10 pt, conference]{ieeeconf}  

\IEEEoverridecommandlockouts                              

\usepackage{graphicx}

\usepackage{subcaption}
\usepackage{graphics} 
\usepackage{amsmath} 
\usepackage{amssymb}  
\usepackage{booktabs}
\usepackage{multirow}
\usepackage{makecell}
\usepackage{cuted}   
\usepackage{capt-of} 

\newif\ifanonymous
\anonymousfalse  

\usepackage[hidelinks]{hyperref}

\usepackage{xcolor}

\ifanonymous
\else
  \IEEEoverridecommandlockouts
\fi

\title{Beyond Human Demonstrations: Diffusion-Based Reinforcement Learning to Generate Data for VLA Training}

\ifanonymous
  \author{Anonymous Authors}
\else
\author{
    Rushuai Yang$^{\ast,1}$,
    Hangxing Wei$^{\ast,3}$,
    Ran Zhang$^{\ast,4}$,
    Zhiyuan Feng$^{5}$,
    Xiaoyu Chen$^{5}$,
    Tong Li, \\
    Chuheng Zhang$^{2,\dagger}$,
    Li Zhao$^{2}$,
    Jiang Bian$^{2}$,
    Xiu Su$^{7}$,
    Yi Chen$^{1,\dagger}$%
    \thanks{This work was done when Rushuai Yang was an intern at Microsoft Research Asia.}%
    \thanks{$^{\ast}$ Equal contribution.}%
    \thanks{$^{1}$ Hong Kong University of Science and Technology, Hong Kong, China.}%
    \thanks{$^{2}$ Microsoft Research Asia, Beijing, China.}%
    \thanks{$^{3}$ Wuhan University, Wuhan, China.}%
    \thanks{$^{4}$ University of Chinese Academy of Sciences, Beijing, China}%
    \thanks{$^{5}$ Tsinghua University, Beijing, China.}%
    \thanks{$^{7}$ Big Data Institute, Central South University, Changsha, China.}%
    \thanks{$^{\dagger}$ Correspondence to: \texttt{chuhengzhang@microsoft.com}, \texttt{yichen@ust.hk}}%
  }
\fi

\makeatletter
\let\@oldmaketitle\@maketitle%
\renewcommand{\@maketitle}{%
  \@oldmaketitle
  \begin{center}
    \captionsetup{type=figure}
    \includegraphics[width=\textwidth]{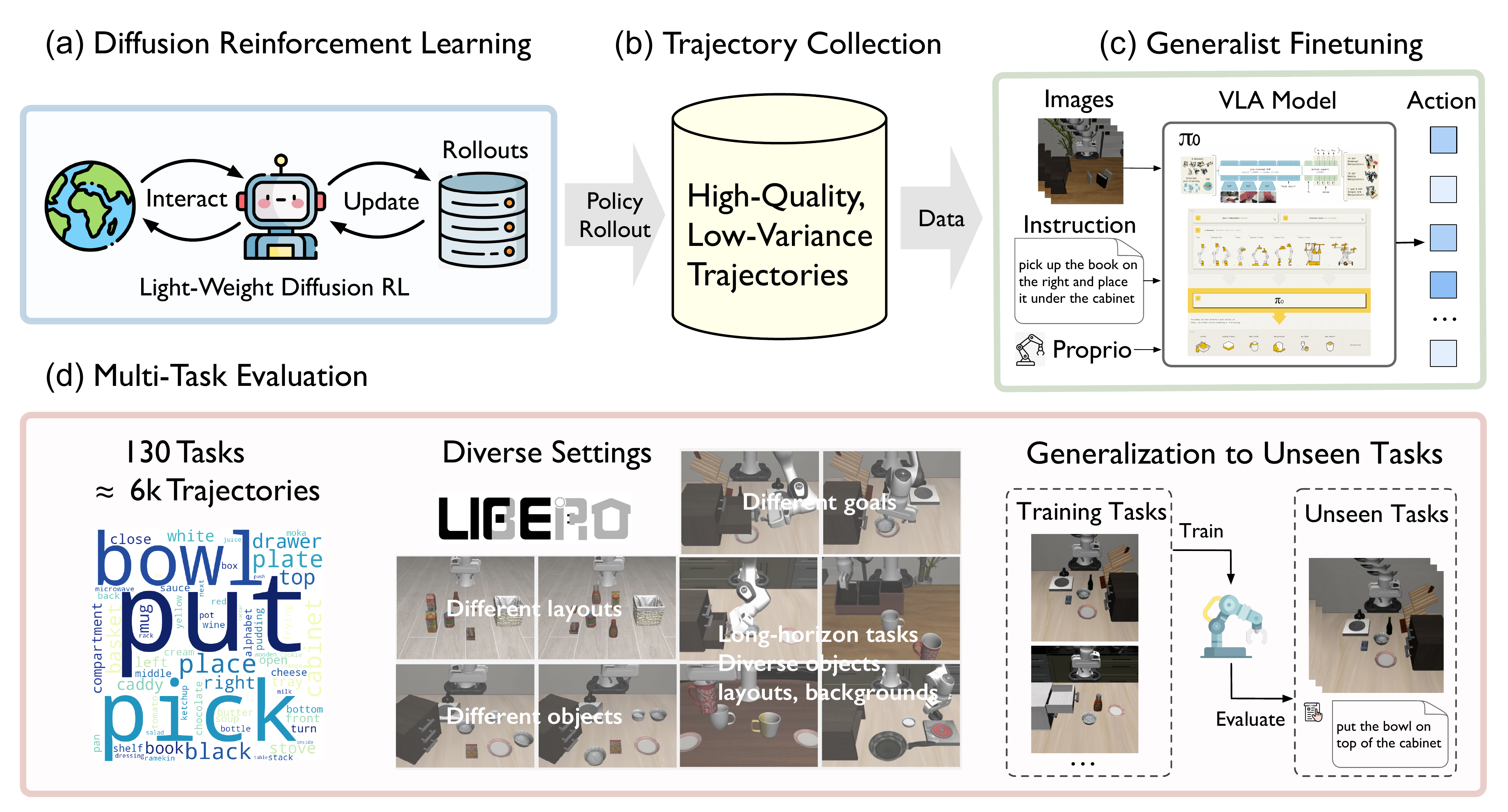}\label{fig:pipeline}
    \caption*{Fig. 1: \textbf{Our diffusion RL-Powered VLA Training Pipeline.}
    \textbf{(a) Diffusion Reinforcement Learning.} 
    A light-weight diffusion policy is optimized for each task via online RL, with only $\,\sim\,$12M parameters in total;
    \textbf{(b) Trajectory Collection.} 
    The optimized policies are used to collect a large dataset of high-quality and low-variance trajectories, containing near-optimal demonstrations for each task;
    \textbf{(c) Generalist Finetuning.} The synthetic dataset is then leveraged to finetune a generalist VLA model, transferring task-specific expertise into a unified policy;
    \textbf{(d) Multi-Task Evaluation.} The VLA model is evaluated on the LIBERO benchmark across diverse tasks and settings, demonstrating that our RL-generated trajectories provide effective supervision for imitation learning and can translate into generalizable performance.}
  \end{center}
  
}
\addtocounter{figure}{1}
\makeatother

\begin{document}
\maketitle
\thispagestyle{empty}
\pagestyle{empty}


\begin{abstract}
Vision-language-action (VLA) models have shown strong generalization across tasks and embodiments; however, their reliance on large-scale human demonstrations limits their scalability owing to the cost and effort of manual data collection. Reinforcement learning (RL) offers a potential alternative to generate demonstrations autonomously, yet conventional RL algorithms often struggle on long-horizon manipulation tasks with sparse rewards. In this paper, we propose a modified diffusion policy optimization algorithm to generate high-quality and low-variance trajectories, which contributes to a diffusion RL-powered VLA training pipeline. Our algorithm benefits from not only the high expressiveness of diffusion models to explore complex and diverse behaviors but also the implicit regularization of the iterative denoising process to yield smooth and consistent demonstrations. We evaluate our approach on the LIBERO benchmark, which includes 130 long-horizon manipulation tasks, and show that the generated trajectories are smoother and more consistent than both human demonstrations and those from standard Gaussian RL policies. Further, training a VLA model exclusively on the diffusion RL-generated data achieves an average success rate of 81.9\%, which outperforms the model trained on human data by +5.3\% and that on Gaussian RL-generated data by +12.6\%. The results highlight our diffusion RL as an effective alternative for generating abundant, high-quality, and low-variance demonstrations for VLA models.
\end{abstract}

\section{Introduction}
\label{sec:introduction}
Vision-language-action (VLA) is a promising model toward general-purpose robots capable of generalizing across a wide array of manipulation tasks~\cite{ma2024survey,sapkota2025vision,zhong2025survey}. However, this paradigm is critically dependent on massive datasets of human demonstrations, such as the Open X-Embodiment dataset~\cite{oxe}. The process of collecting this data via manual teleoperation is notoriously expensive and labor-intensive. The reliance on manual data collection fundamentally caps the scalability of VLA models, presenting a major bottleneck to further progress.

Reinforcement learning (RL) has emerged as a powerful paradigm for enabling robots to acquire sophisticated physical skills directly through environmental interaction. 
The fundamental strength of reinforcement learning stems from its trial-and-error process: by optimizing for a reward signal, an agent can autonomously discover highly effective and efficient strategies that often surpass what can be learned by simply mimicking human demonstrations. 
However, a significant limitation of this approach is that the resulting policies are often highly specialized. 
A policy trained to excel under one specific set of conditions typically struggles to adapt or generalize its skills when faced with new task variations or different environmental setups~\cite{sancaktar2022curious}. 
However, making a general RL algorithm effective enough to generate high-quality data across diverse, complex manipulation tasks is challenging. 
The long-horizon, sparse-reward tasks prevalent in benchmarks like LIBERO~\cite{libero} expose critical weaknesses of conventional RL algorithms, often leading to unstable learning process or high-variance and suboptimal trajectories~\cite{zhou2025spire}.
\begin{figure}[t]
    \centering
    \includegraphics[width=\columnwidth]{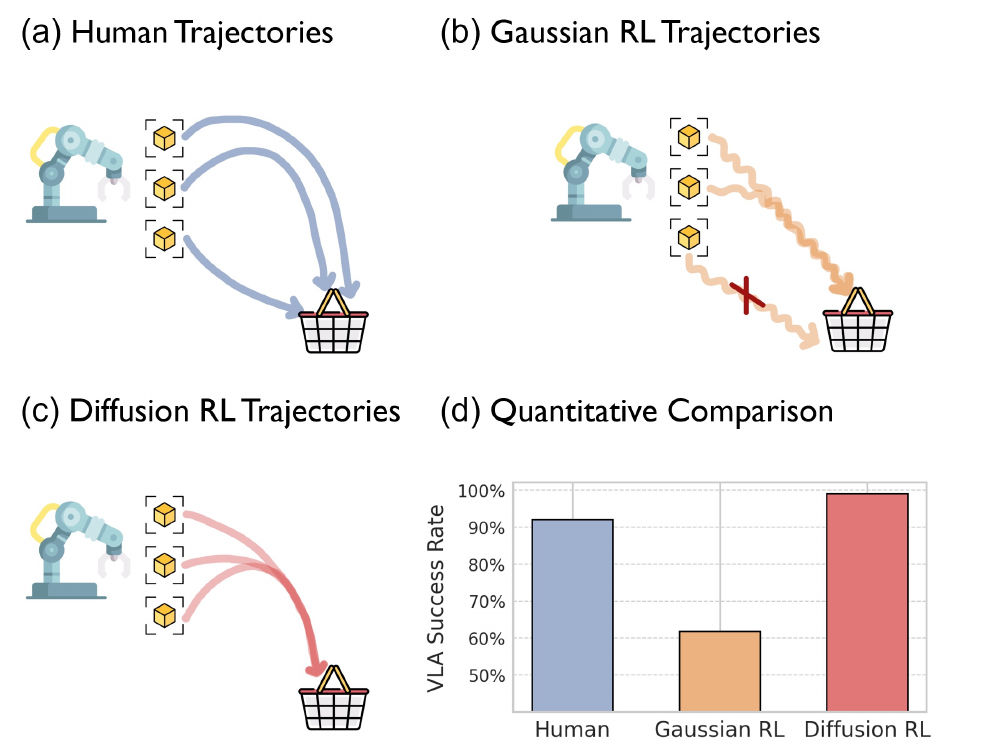}
    \caption{\textbf{Our diffusion RL method generates a superior data distribution for VLA training.} The figure contrasts three types of generated trajectories. 
(a) Human demonstrations, while successful, are inconsistent and exhibit high-variance, multi-modal behaviors;
(b) A conventional Gaussian RL policy produces jittery, suboptimal trajectories that result in poor task success;
(c) In contrast, our method consistently generates high-quality, low-variance trajectories that are both smooth and optimal;
(d) This superior data quality directly translates to downstream performance, as the VLA model trained on our data significantly outperforms those trained on the others. Quantitative results can be found in Sec.~\ref{sec: Data quality Analysis}.
}\label{fig:data_quality_comparison}
\end{figure}
To this end, we propose a general framework that utilizes a modified diffusion policy optimization algorithm for diffusion RL-powered data generation. 
We find that diffusion offer a superior alternative for this problem. 
First, diffusion policy provide good expressiveness to fit complex expert distribution.
Compared with Gaussian RL, diffusion-based RL provides more space for RL exploration when interacting with environment. 
Second, the inherent structure of the iterative denoising process acts as a powerful implicit regularizer on the action space, The model is trained to predict the noise for the entire action chunk at every step of the denoising process. This forces the model to learn the underlying structure of smooth, physically plausible motions. A single, jerky movement in the final action would require a very specific and complex sequence of denoising steps, which is less likely to be learned than a smooth, coherent refinement process. This naturally encourages the generation of temporally smooth, low-variance motion. We further enhance this process with a stabilized fine-tuning regimen, incorporating modifications to the architecture and training strategies to ensure robust performance across the 130 challenging tasks in LIBERO. This property allows our RL agent to explore more effectively and converge to near-optimal and low-variance policies.

Our experiments yield a clear message: 
A VLA model trained exclusively on our RL-generated data consistently and significantly surpasses the ones trained on human data and Gaussian RL, both on in-distribution tasks and in challenging OOD generalization. A quantitative analysis reveals the mechanism behind this success: Our generated trajectories are smoother and less variable, providing a more stable learning signal for VLA training.
Our contributions are threefold:
\begin{itemize}
\item \textbf{A diffusion RL-powered VLA training pipeline} for autonomously generating high-quality and low-variance data for VLA training, including validated effective modifications on the model architecture and training strategies.
\item \textbf{Compelling empirical evidence} on the 130 complex manipulation tasks of the LIBERO benchmark shows that our synthetic data provides superior training signal to human demonstrations, significantly improving both the in-distribution success rates and out-of-distribution generalization of VLA models.
\item \textbf{An in-depth quantitative analysis} that relates trajectory-level properties (e.g., efficiency, smoothness, and consistency) with the performance of fine-tuned VLA, providing a clear explanation for why optimized data is more effective.
\end{itemize}

\begin{figure*}[!t]
    \centering
    \includegraphics[width=\textwidth]{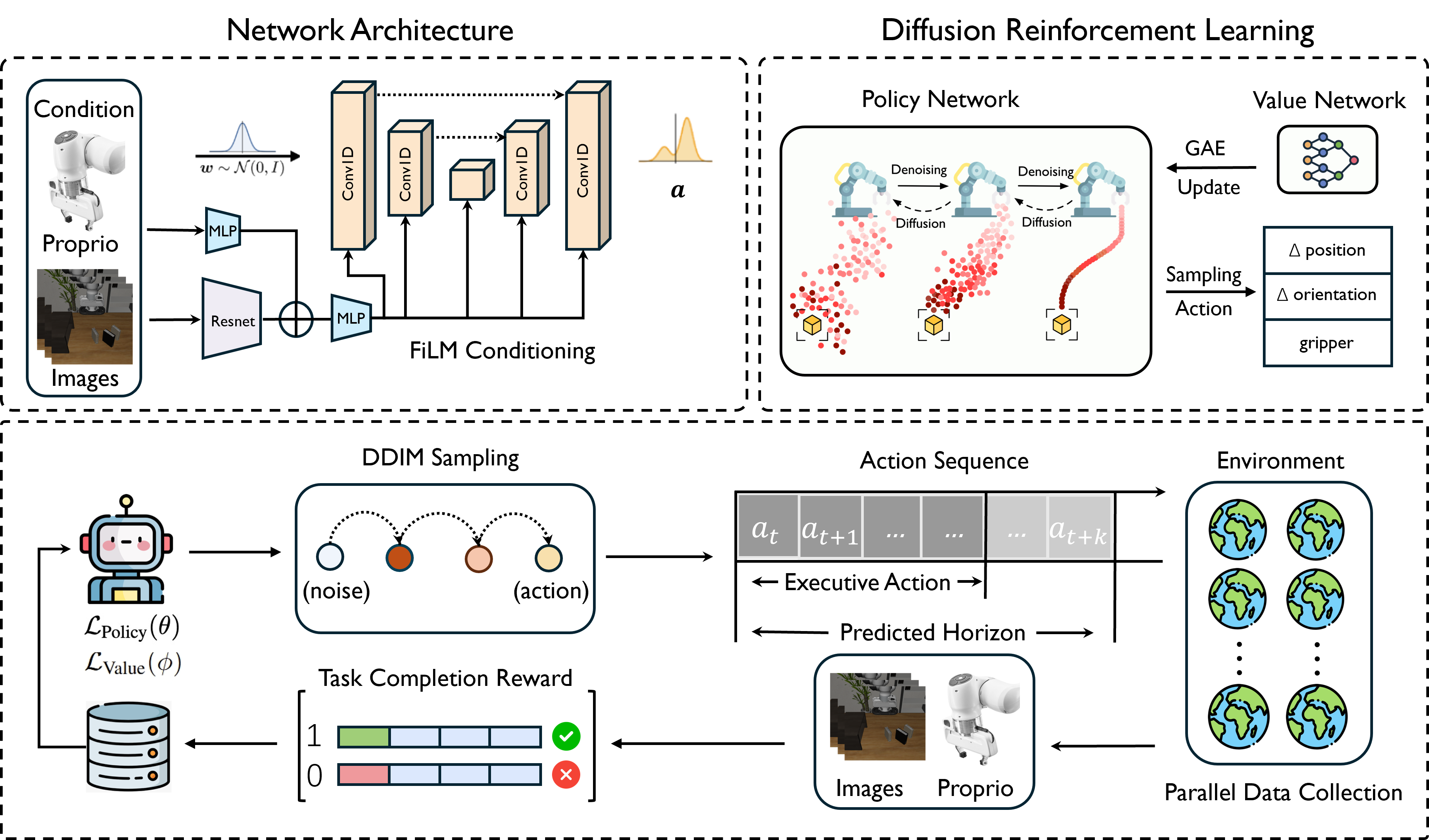}
    \caption{\textbf{Overview of our diffusion RL framework.} (Top-left) The policy network architecture integrates proprioceptive and visual observations via a ResNet backbone and FiLM conditioning to parameterize the denoising process. (Top-right) The policy network samples actions through iterative diffusion denoising, while the value network estimates the value of these actions to guide the policy network’s updates. (Bottom) The training loop executes diffusion-based action sampling within parallel environment rollouts, collecting task-completion rewards to update the networks and improve policy performance.}
    \label{fig:RL_algorithm}
\end{figure*}

\section{Related Work}
\label{sec:related_work}

\subsection{Data Generation for VLA Models}
While many recent VLA models have demonstrated impressive capabilities, their performance often relies on the availability of large-scale datasets. 
Open-source datasets, such as Open X-Embodiment~\cite{oxe} and LIBERO~\cite{libero}, provide a strong foundation for the field, with most of this data being collected through human teleoperation. 
A key challenge is that scaling this data collection process requires significant human labor. On another front, some model-based methods generate trajectories for manipulation tasks by solving optimization problems~\cite{mujoco}, but this often requires careful, task-specific algorithmic design. 
Seeking a more scalable and autonomous approach, we explore using RL for data generation. 
The ability of RL to discover near-optimal behaviors through online interaction presents a promising path towards generating high-quality data.

\subsection{Reinforcement Learning in Robotic Foundation Models}
Recent works have started to explore how RL algorithms can be applied to enhance robotic foundation models.
One key challenge is how to train the RL agents effectively on long-horizon robotic tasks with sparse rewards.
To overcome this challenge, researchers incorporate RL with different strategies such as leveraging foundation priors~\cite{ye2023reinforcement}, refining policies via distillation~\cite{julg2025refined}, leveraging human-in-the-loop~\cite{luo2025precise}, training offline~\cite{shah2022offline}, or interactive RL methods~\cite{tan2025interactive}.
Another line of research adopts RL to fine-tune the large vision–language backbones directly, demonstrating improvements in decision-making capabilities~\cite{zhao2025unlocking,guo2025improving,zhang2024grape,zhai2024fine}.
However, directly finetuning VLA via online RL is computationally costly due to the sample inefficiency of RL.
Hence, another promising route is to employ RL as a tool for scalable data generation. 
For example, Liu et al.~\cite{liu2025can} and Xu et al.~\cite{xu2024rldg} show that RL-generated trajectories can reduce the noise and inconsistency usually found in human demonstrations. 
This provides a high-quality supervision signal for VLA training and can be easily combined with other sources of data, such as curated human collected data~\cite{o2024open} and world model generated data~\cite{zhou2024genesis} and integrated into the standard VLA training pipeline.
Built upon this line of work, our approach introduces a modified diffusion-based RL algorithm that can robustly generates high-quality low-variance trajectories, providing a scalable alternative/complement to human-collected data for training robotic foundation models.

\graphicspath{{figures/}} 

\section{Method}
\label{sec:method}

Our objective is to develop a well-designed pipeline for generating high-quality synthetic data via RL to train VLA models, thereby reducing the dependency on costly and inconsistent human demonstrations. Our end-to-end process involves three key stages: \textbf{(1) Training expert RL policies}, where we use a two-phase paradigm of BC warm-start followed by online RL fine-tuning to train an expert diffusion-based policy for each task. 
(2) \textbf{Generating synthetic dataset}, where we deploy the converged expert policies to autonomously collect a dataset of optimal trajectories.
(3) \textbf{Training the generalist VLA model}, where this synthetic dataset is used to fine-tune a generalist VLA model. The core of our method is the use of a conditional diffusion model as the policy architecture, which is uniquely suited to overcome the distinct challenges present in each phase.
The illustration of the pipeline can be found in Fig.~1.
\subsection{Preliminaries: Diffusion Policies for Manipulation}
Directly applying vanilla RL to complex manipulation tasks is challenging due to long horizons and sparse rewards. A common strategy to mitigate this is to warm-start a policy on a small amount of human demonstrations. However, human demonstrations are inherently multimodal, reflecting diverse operational strategies for the same task~\cite{diffusionpolicy}. Standard policies assuming a unimodal Gaussian distribution struggle to learn from such data, often averaging over distinct modes and resulting in a poor starting point for RL fine-tuning. To effectively capture this rich knowledge for the next RL stage, we utilize the high expressive power of a diffusion policy.
\subsection{Phase 1: Multimodal Behavior Cloning as Warm-Start}
Specifically, a diffusion policy $\pi_\theta$ models the distribution over an action chunk $a_0$ by learning to reverse a diffusion process that gradually corrupts the action with Gaussian noise over $K$ steps. 
The core of the model is a neural network $\epsilon_\theta(a_k, s_t, k)$ that predicts the noise added to a clean action $a_0$ to produce a noisy action $a_k$ at step $k$, conditioned on the state $s_t$. An action is generated by starting with pure noise $a_K \sim \mathcal{N}(0, \mathbf{I})$ and iteratively applying the learned denoising function to produce a clean action $a_0$. 
In the first phase, we train the diffusion policy to mimic a few multi-modal human demonstrations $\mathcal{D}_{\text{human}}$. The noise prediction network $\epsilon_\theta$ is optimized with the simplified objective $\mathcal{L}_{\text{BC}}
(\theta)$~\cite{ho2020denoising}:
\begin{equation}
     \mathbb{E}_{\substack{k\sim\mathrm{Unif}\{1..K\},\\ (s_t,a_0)\sim\mathcal{D}_{\text{human}},\\ \epsilon\sim\mathcal{N}(0,\mathbf{I})}} \Big[ 
    \big\| \epsilon - \epsilon_\theta(\sqrt{\bar{\alpha}_k} a_0 + \sqrt{1 - \bar{\alpha}_k}\epsilon, s_t, k) \big\|^2 \Big],
    \label{eq:bc_loss}
\end{equation}
where $\bar{\alpha}_k$ is a pre-defined noise schedule coefficient. Optimizing Eq.~\eqref{eq:bc_loss} allows the policy to capture the complex, multi-modal distribution of human behaviors, providing a high-quality starting point for the subsequent RL phase.

\subsection{Phase 2: Online Reinforcement Learning using PPO}
In the second phase, we train the warmed-up policy using online RL to maximize the expected cumulative reward. A key challenge is that $\pi_\theta(a_0|s_t)$ is intractable for a diffusion policy. We thus adopt the insight from recent work~\cite{diffusionrl, ren2024diffusion, diffusionpolicy} which treats the denoising process as a sub-trajectory of decisions. This allows us to apply policy gradient methods like PPO by leveraging the tractable, single-step transition likelihood $p_\theta(a_{k-1}|a_k, s_t)$.

The value network $V_\phi(s)$, is trained to minimize the temporal-difference error:
\begin{equation}
    \mathcal{L}_{\text{Value}}(\phi) = \mathbb{E}_{t} \left[ (V_\phi(s_t) - \hat{R}_t)^2 \right],
    \label{eq:critic_loss}
\end{equation}
where $\hat{R}_t$ are the target values, typically computed using generalized advantage estimation (GAE)~\cite{schulman2015high}:
\begin{equation}
    \hat{A}_t = \sum_{l=0}^\infty (\gamma\lambda)^l \delta_{t+l}, \quad \text{where} \quad \delta_t = r_t + \gamma V_\phi(s_{t+1}) - V_\phi(s_t).
\end{equation}
The diffusion policy $\pi_\theta$, is updated by maximizing the PPO-clip objective~\cite{schulman2017proximal}. The gradient is computed over the sequence of denoising steps for each action generated during environment interaction $\mathcal{L}_{\text{Policy}}(\theta)$:
\begin{equation}
\mathbb{E}_{t, k} \left[ \min(r_{t,k}(\theta)\hat{A}_t, \text{clip}(r_{t,k}(\theta), 1-\epsilon, 1+\epsilon)\hat{A}_t) \right],
    \label{eq:actor_loss}
\end{equation}
where $\hat{A}_t$ is the advantage estimate from the critic, and $r_{t,k}(\theta)$ is the likelihood ratio for a single denoising step:
\begin{equation}
    r_{t,k}(\theta) = \frac{p_\theta(a_{k-1}|a_k, s_t)}{p_{\theta_{\text{old}}}(a_{k-1}|a_k, s_t)}.
\end{equation}

\subsection{Stabilizing the Reinforcement Learning Process}
Directly optimizing the objectives in Eq.~\eqref{eq:critic_loss} and \eqref{eq:actor_loss} with diffusion policy presents significant stability challenges. 
To make the process practical and robust, we propose key changes to the architecture, sampling strategy, and training regimen within the pipeline. The full ablation can be found in Sec.~\ref{sec:ablation_study}.

\subsubsection{Architectural Design for Sample Efficiency and Stability}
While some prior works \cite{ren2024diffusion} have utilized Vision Transformer (ViT) and MLP architectures, our empirical investigation revealed this combination is suboptimal for the dual challenges of LIBERO: 
supervised learning from limited, multi-modal human data and 
reinforcement learning on long-horizon tasks. We found that a ResNet backbone paired with a U-Net decoder is a superior solution. The strong inductive bias of ResNet results in higher sample efficiency in the low-data regime, while the U-Net is crucial for effectively modeling the multi-modal nature of human demonstrations, providing a robust warm-start where a simple MLP fails. 
We further enhance this architecture by integrating proprioceptive state information via a FiLM mechanism for a more stable conditioning signal during RL updates.

\subsubsection{Efficient Action Sampling with Optimized Schedulers}
A significant bottleneck in diffusion RL is the slow iterative nature of the standard denoising diffusion probabilistic models (DDPM) sampler~\cite{ho2020denoising}. It is computationally expensive and introduces high variance into the sampled actions, which can destabilize the critic updates. To overcome this, we switch to a faster, deterministic denoising diffusion implicit models (DDIM) sampler~\cite{song2020denoising}. During reinforcement learning, we use only 5 denoising steps to generate each action. 
This leads to a more stable and reliable policy gradient, accelerates data collection, and reduces action variance.

\begin{figure*}[!t]
    \centering
    \includegraphics[width=\linewidth]{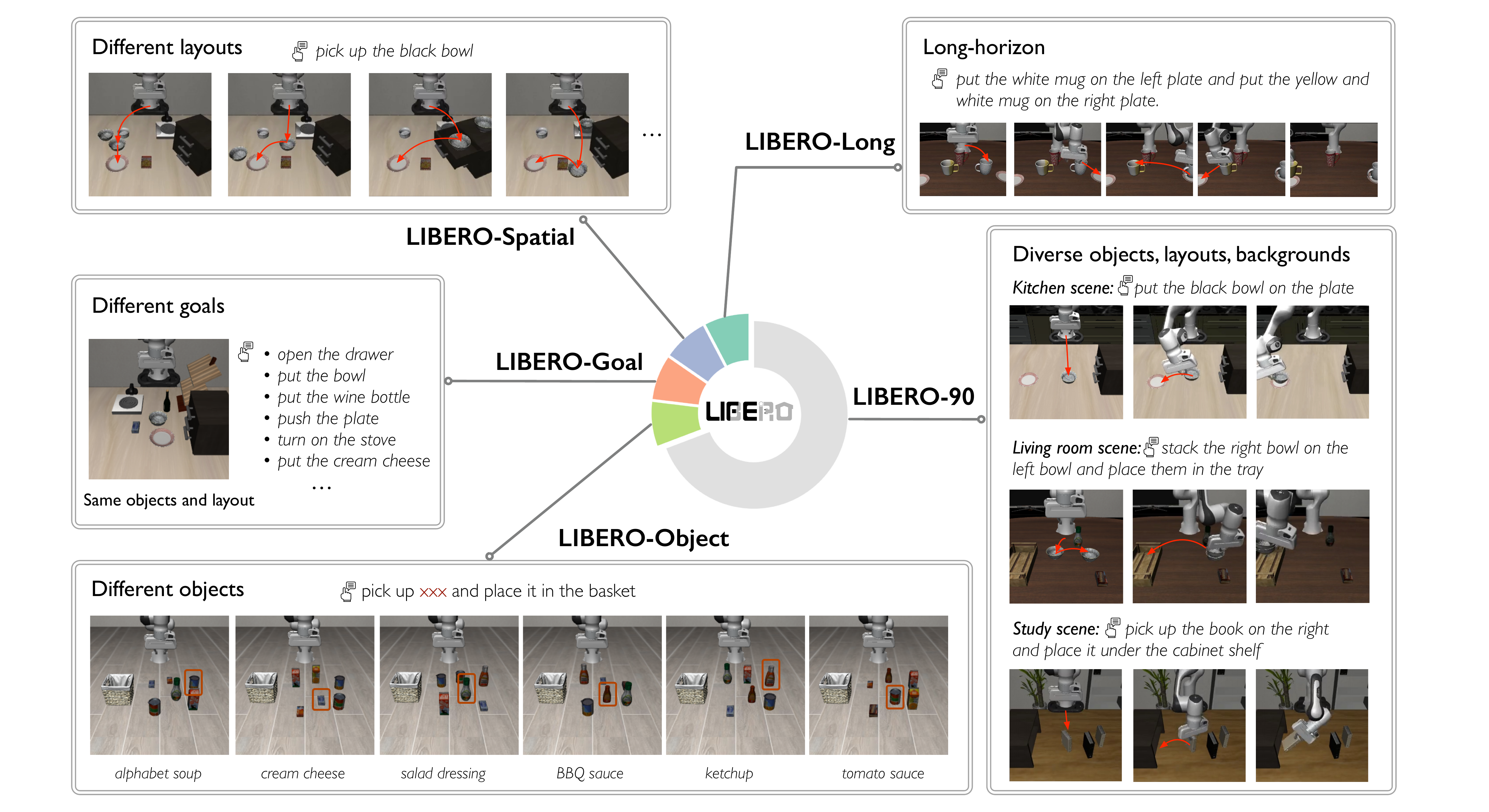} 
    \caption{\textbf{Overview of the LIBERO benchmark, a suite of 130 robot manipulation tasks.} It includes subsets targeting different challenges: varying layouts (LIBERO-Spatial), goals (LIBERO-Goal), objects (LIBERO-Object), long-horizon tasks (LIBERO-Long), and diverse scenes (LIBERO-90). These suites provide a comprehensive testbed for benchmarking diverse manipulation skills.}
    \vspace{-2mm}
    \label{fig:experiment setup}
\end{figure*}

\subsubsection{Training Regimen for Stable Exploration}
Fine-tuning a high-capacity diffusion policy requires a careful balance between preserving the rich prior learned from BC and exploring new potential actions. To this end, we explore a set of more effective training regimens:
\begin{itemize}
    \item \textbf{Annealed Learning Rate:} We employ a cosine annealing learning rate schedule \cite{loshchilov2017sgdr}. It begins with a larger learning rate to encourage exploration beyond the initial human demonstrations and then gradually decays to stabilize convergence and retain the powerful priors learned from the BC phase.
    \item \textbf{Diverse Experience Replay to Prevent Mode Collapse:} 
    During the fine-tuning process, we found that a key factor for training stability is ensuring a sufficiently large and diverse batch of experiences for each PPO iteration. This is particularly critical for our high-capacity diffusion policy. Due to its expressiveness, if the policy is updated on a small batch of correlated data (e.g., from similar trajectories), it can quickly overfit to the bias within these trajectories. This causes its learned multi-modal distribution to collapse into a single but suboptimal behavior. A collapsed policy is used for rollout in the subsequent iterations and yields vicious samples which may result in performance degradation.
    We utilize a large number of parallel environments to efficiently populate the replay buffer without sacrificing training speed.
\end{itemize}
\subsection{VLA Training on Generated Data}
Upon convergence, the expert diffusion-RL policies are used to generate the synthetic dataset $\mathcal{D}_{\text{RL}} = \{ (o_t, a_t, l)_i \}_{i=1}^N$. We then train the VLA model, $\Pi_{\text{VLA}}$, by minimizing a general behavioral cloning objective. This is achieved by maximizing the log-likelihood of the expert actions from our dataset, which is equivalent to minimizing the following negative log-likelihood loss:
\begin{equation}
    \mathcal{L}_{\text{VLA}} = - \mathbb{E}_{(o_t, a_t, l) \in \mathcal{D}_{\text{RL}}} 
    \left[ \log \Pi_{\text{VLA}}(a_t | o_t, l) \right].
    \label{eq:vla_loss}
\end{equation}
This formulation is general and accommodates different policy architectures. 
For a policy with a continuous action space like $\pi_0$~\cite{pi}, this objective is commonly implemented by minimizing the mean squared error between the predicted and target actions, or by optimizing a diffusion model's noise prediction loss as in Eq.~\eqref{eq:bc_loss}. We use this procedure to fine-tune pre-trained VLA architectures, ensuring a fair comparison across different data sources by keeping all VLA training hyperparameters identical.

\graphicspath{{figures/}}

\section{EXPERIMENTS}
\label{sec:experiments}

\subsection{Experimental Setup}

\begin{table*}[th]
\caption{\textbf{In-distribution success rates (SR).} VLA models trained on our diffusion RL-generated data outperform other baselines on average. We run 50 evaluations per task on all 90 (for LIBERO-90) and 10 (for others) tasks in each set.}
\label{table: Evaluation on 5 LIBERO task suites}
\centering
\begin{tabular}{lcccccc}
\toprule
Data Source
 & \makecell{\textbf{LIBERO-Spatial}\\ SR (\%)} 
 & \makecell{\textbf{LIBERO-Object}\\ SR (\%)} 
 & \makecell{\textbf{LIBERO-Goal}\\ SR (\%)} 
 & \makecell{\textbf{LIBERO-Long}\\ SR (\%)} 
 & \makecell{\textbf{LIBERO-90}\\ SR (\%)} 
 & \makecell{\textbf{Average}\\ SR (\%)} \\
\midrule
Human data & 73.40 & 92.00 & 83.20 & \textbf{66.00} & 68.60 &  76.64        \\
Gaussian RL data    & 80.40 & 61.80 & 82.40 & 47.00 & 75.00 &  69.32  
\\
Diffusion RL data (Ours)     & \textbf{83.40} & \textbf{99.00} & \textbf{83.80} & 63.00 & \textbf{80.49} &  \textbf{81.94}  \\
\bottomrule
\end{tabular}%
\end{table*}

\begin{table*}[th]
\centering
\caption{\textbf{Out-of-distribution success rate (SR).} VLA models are pretrained on LIBERO-90 with different data sources and evaluated zero-shot on entirely unseen task suites, spanning 30 unseen tasks, objects, and goals.}
\label{tab:ood_eval}
\begin{tabular}{lcccc}
\toprule
Data Source
 & \makecell{\textbf{LIBERO-Long}\\ SR (\%)} 
 & \makecell{\textbf{LIBERO-Goal}\\ SR (\%)} 
 & \makecell{\textbf{LIBERO-Object}\\ SR (\%)} 
 & \makecell{\textbf{Average}\\ SR (\%)} \\
\midrule
Human data   & 0.00 & 0.00 & 4.40 & 1.47 \\
Diffusion RL data     & 1.40 & 0.00 & 4.80 & 2.06 \\
Human data + Diffusion RL data     & \textbf{4.60} & \textbf{5.60} & \textbf{5.40} & \textbf{5.20} \\
\bottomrule
\end{tabular}
\end{table*}

\begin{figure*}[!t]
  \centering
  \begin{subfigure}[t]{0.24\textwidth}
    \centering
    \includegraphics[width=\linewidth]{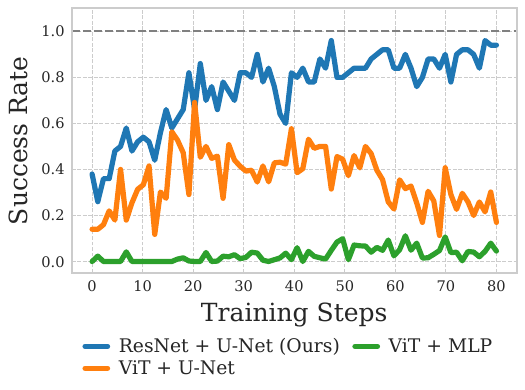}
    \caption{Policy architecture}
    \label{fig:abl-arch}
  \end{subfigure}\hfill
  \begin{subfigure}[t]{0.24\textwidth}
    \centering
    \includegraphics[width=\linewidth]{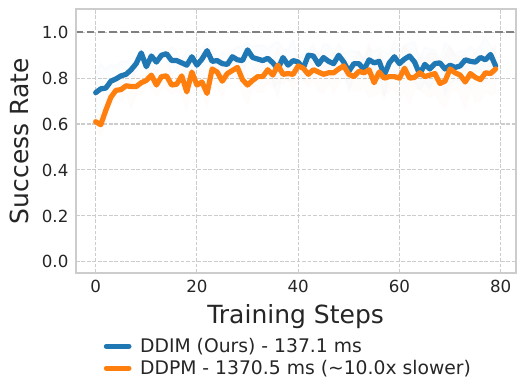}
    \caption{DDPM vs. DDIM}
    \label{fig:abl-ddpmddim}
  \end{subfigure}\hfill
  \begin{subfigure}[t]{0.24\textwidth}
    \centering
    \includegraphics[width=\linewidth]{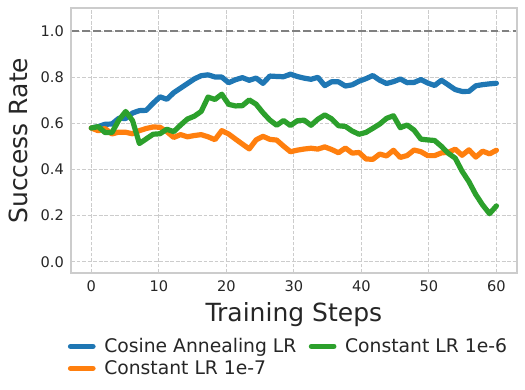}
    \caption{Learning rate}
    \label{fig:abl-lr}
  \end{subfigure}\hfill
  \begin{subfigure}[t]{0.24\textwidth}
    \centering
    \includegraphics[width=\linewidth]{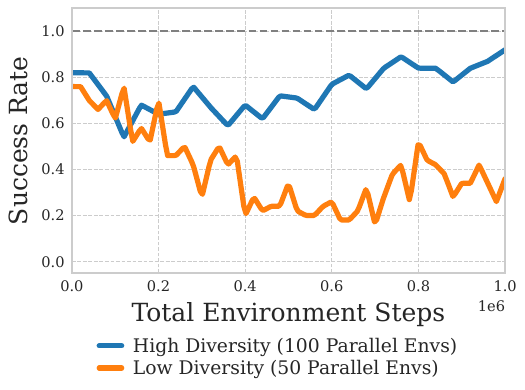}
    \caption{Replay buffer diversity}
    \label{fig:abl-replay}
  \end{subfigure}

  \caption{\textbf{Ablation study} across architecture, diffusion sampling, learning rate, and data diversity in RL on LIBERO tasks.}
  \vspace{-2mm}
  \label{fig:abl-1x4}
\end{figure*}

\paragraph{The LIBERO benchmark}
We validate our approach on the comprehensive LIBERO benchmark~\cite{libero}, a suite of 130 manipulation tasks designed for studying knowledge transfer. LIBERO is exceptionally challenging for RL algorithms for two primary reasons. First, its tasks are often \emph{long-horizon with sparse rewards}, requiring agents to execute long sequences of actions before receiving any success signal, which makes exploration difficult. Second, it demands the transfer of both declarative knowledge (e.g., recognizing new objects in \texttt{LIBERO-Object}) and procedural knowledge (e.g., learning new behaviors in \texttt{LIBERO-Goal}), 
which is a significant challenge.
The benchmark is split into suites like \texttt{LIBERO-Spatial}, \texttt{LIBERO-Object}, \texttt{LIBERO-Goal}, and the large-scale \texttt{LIBERO-100} (composed of \texttt{LIBERO-90} and \texttt{LIBERO-Long}), allowing for controlled studies of generalization.

\paragraph{Data Generation Methods for Comparison}
To evaluate the quality of the generated data, we compare VLA models trained on four distinct datasets:
\begin{itemize}
    \item \textbf{Human Data:} The original dataset provided by LIBERO~\cite{libero}, containing 50 human teleoperated demonstrations for each task. 
    \item \textbf{Gaussian RL Data:} 
    We follow prior work~\cite{xu2024rldg} in the field that leverages Gaussian reinforcement learning agents to generate synthetic trajectories for comparison with or augmentation of human-collected data, ensuring consistency with established practices. We follow a standard PPO under the using the same architecture and training procedure as our main agent.
\item \textbf{Diffusion RL Data (Ours):} A dataset of 50 trajectories per task generated by our converged, stabilized diffusion RL agent.
    \item \textbf{Human+RL Data:} A mixed dataset combining both human and our diffusion RL data, used specifically for OOD analysis. 
\end{itemize}
We follow the data preprocessing as mentioned from OpenVLA~\cite{o2024open} to keep a fair comparison of these baselines.

\paragraph{Evaluation Protocol}
The primary metric is the task success rate, averaged over 50 evaluation episodes per task suite. In each trial, every task is randomly initialized with different states, including variations in target positions and the placement of surrounding objects. For OOD evaluation, models are trained exclusively on the \texttt{LIBERO-90} suite and then evaluated zero-shot on unseen tasks from other suites.

\subsection{Main Results}

We compare the performance of VLA models trained on our diffusion RL data versus those trained on human and Gaussian RL data. Our findings show that our diffusion-based synthetic data is a superior training signal for in-distribution tasks.

\subsubsection{Diffusion RL Data Excels for In-Distribution Performance}
As shown in Table~\ref{table: Evaluation on 5 LIBERO task suites}, the choice of data generator is critical. A VLA model trained on data from a standard Gaussian PPO agent performs poorly, achieving an average success rate of only 69.32\%. This is significantly worse than using human data and is attributed to the Gaussian policy's instability on the complex LIBERO tasks, leading to the generation of suboptimal and failed trajectories. 

In contrast, VLA models trained exclusively on our diffusion RL-generated data consistently outperform those trained on human data across most task suites. For the $\pi_0$ model, this results in a significant +5.3\% average improvement over human data and a massive +12.6\% improvement over Gaussian RL data. The largest gains are seen in the \texttt{LIBERO-Object} and \texttt{LIBERO-Spatial} suites, suggesting that the consistency and optimality of diffusion-RL trajectories provide a clearer learning signal for tasks involving precise object interaction and spatial reasoning.

\subsubsection{Synergy of RL and Human Data for OOD Generalization}
To evaluate the generalization capabilities of different data sources, we conduct a challenging OOD experiment. We pre-train the $\pi_0$ VLA model exclusively on the \texttt{LIBERO-90} task suite using three different data sources, and then evaluate its zero-shot performance on completely unseen task suites: \texttt{LIBERO-Long}, \texttt{LIBERO-Goal}, and \texttt{LIBERO-Object}.

As shown in Table~\ref{tab:ood_eval}, models pre-trained on a single data source exhibit extremely limited generalization. The model trained on Human-only data fails to solve any tasks in the \texttt{LIBERO-Long} and \texttt{LIBERO-Goal} suites, achieving a meager 1.47\% average success rate. Using our RL-generated data alone provides an improvement to 2.06\%, but still struggles with transferring procedural knowledge to the unseen \texttt{LIBERO-Goal} tasks.
The most significant result is the powerful synergy observed when combining the two data modalities. While OOD generalization on LIBERO remains a difficult open problem~\cite{libero,pi, openvla}, this result is more than double the performance of either single data source, demonstrating a clear complementary benefit. We conclude that the diversity and multi-modal exploration strategies from human data, when combined with the consistency and optimality of RL-generated trajectories, provide a richer and more robust pre-training signal for downstream generalization.
\subsection{Ablation Studies}
\label{sec:ablation_study}

We conducted a series of ablation studies to validate our design choices (Fig.~\ref{fig:abl-1x4}). Our experiments confirm that a ResNet+U-Net architecture is critical, as the ResNet's sample efficiency and the U-Net's ability to model multi-modal data prevent the policy collapse seen with ViT+MLP alternatives. We also show that using a DDIM sampler (5 steps) provides a nearly 10x speedup in action generation over a standard DDPM sampler with comparable or better performance. Furthermore, a cosine annealing learning rate schedule is vital for stability, outperforming fixed learning rates that either collapse or learn too slowly. Finally, we demonstrate that using a high degree of data diversity from parallel environments is essential for preventing mode collapse and achieving a stable, high-performance final policy.


\begin{figure}[t]
    \centering
    \begin{subfigure}[b]{0.48\columnwidth}
        \centering
        \includegraphics[width=\textwidth]{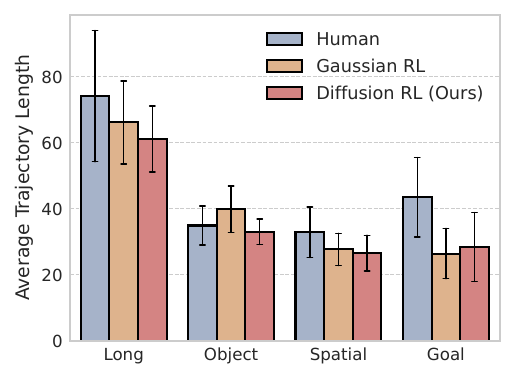}
        \caption{Task Efficiency}
        \label{fig:quality-charts-a}
    \end{subfigure}
    \hfill 
    \begin{subfigure}[b]{0.48\columnwidth}
        \centering
        \includegraphics[width=\textwidth]{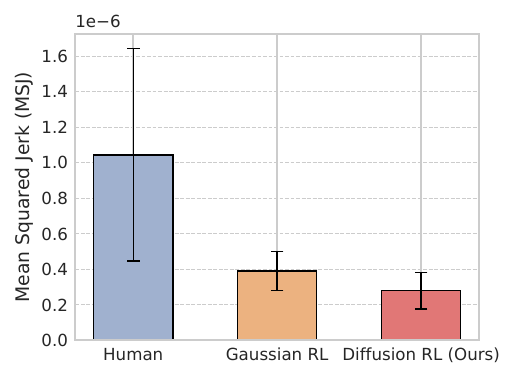}
        \caption{Trajectory Smoothness}
        \label{fig:quality-charts-b}
    \end{subfigure}
    \caption{\textbf{Quantitative comparison of trajectory quality.} (a) Our diffusion RL algorithm generates the most efficient policies, requiring the shortest average trajectory length to complete tasks; (b) It also produces the smoothest motions, exhibiting the lowest Mean Squared Jerk~\cite{balasubramanian2015analysis} compared to both human and Gaussian RL data.}
    \label{fig:quality-charts}
\end{figure}

\begin{figure}[t]
    \centering
    \includegraphics[width=0.85\columnwidth]{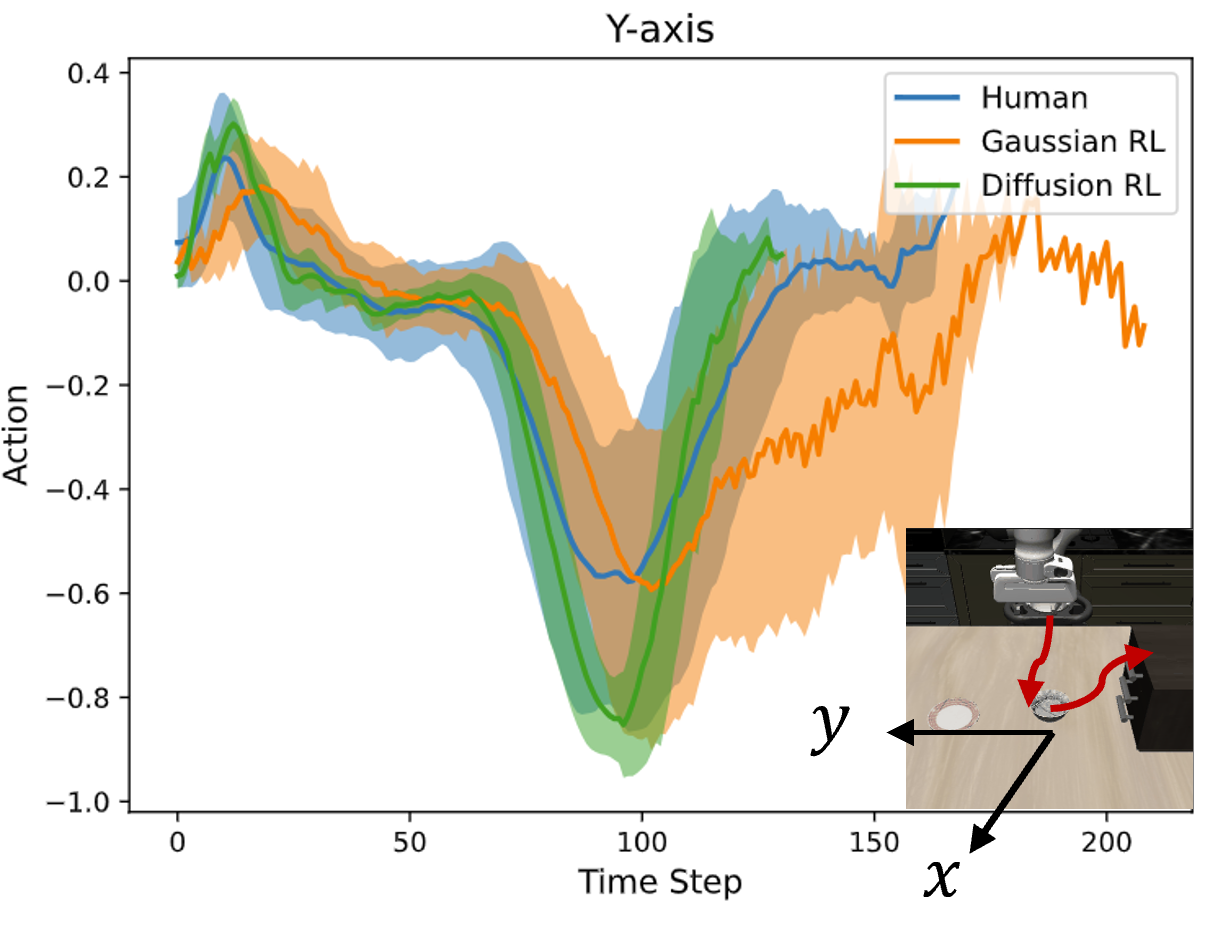}
    \caption{
        \textbf{Comparison of action consistency along the y-axis over time.}
        The solid lines represent the mean action, while the shaded regions denote the standard deviation across 50 successful trajectories for the task ``put the black bowl on top of the cabinet". 
        Our diffusion RL policy exhibits the lowest variance, indicating a consistent, near-deterministic behavior. 
        In contrast, human data shows high variance due to its multi-modality, and the Gaussian RL policy exhibits unstable motion and high-frequency jitter especially at the end of the trajectory.
    }
    \label{fig:quality-y-action}
\end{figure}

\section{Data Quality Analysis}
\label{sec: Data quality Analysis}
We quantitatively compare the properties of datasets from three sources: human teleoperation, a conventional Gaussian RL policy, and our proposed diffusion RL policy. The analysis confirms that our method produces data that is significantly more efficient, smooth, and consistent.

\paragraph{Task Efficiency and Action Redundancy}
\textbf{Our diffusion RL agent learns the most efficient and direct policies.} We first measure action redundancy by counting ``no-op" actions—timesteps where the arm's velocity is near-zero and the gripper state remains unchanged, indicating operator hesitation. Across 4500 trajectories from the \texttt{LIBERO-90} suite~\cite{libero}, human demonstrations contained 3007 no-op instances, whereas both our diffusion RL and the Gaussian RL methods produced zero. Furthermore, as shown in Fig.~\ref{fig:quality-charts}a, our diffusion RL agent completes tasks with the shortest average trajectory length, particularly on long-horizon tasks. This demonstrates that our agent discovers more direct and purposeful solutions, an efficiency that stems from the RL objective which optimizes for cumulative reward, naturally penalizing the hesitant or inefficient motions present in human data.

\paragraph{Trajectory Smoothness}
\textbf{Our method generates the smoothest trajectories, avoiding the jitter common in conventional RL.} We measure smoothness by calculating the Mean Squared Jerk~\cite{balasubramanian2015analysis} of the end-effector path, where lower values are better. The results, visualized in Fig.~\ref{fig:quality-charts}b, show that trajectories from the Gaussian RL policy exhibit significantly higher jerk than both human and diffusion RL data. Our diffusion RL agent produces trajectories with the lowest jerk, indicating a superior level of stability. This is because the iterative denoising process of the diffusion model acts as an implicit regularizer, inherently favoring temporally coherent action sequences over the high-frequency, jittery actions that a standard Gaussian policy can produce.

\paragraph{Action Consistency (Low Variance)}
\textbf{The policies generated by our diffusion RL agent are consistent and near-deterministic.} 
We analyze this by measuring the variance of actions across 50 successful trajectories for the same task, with the results for the y-axis control signal visualized in Fig.~\ref{fig:quality-y-action}. The wide shaded region for the human data indicates high variance, reflecting the diverse, multi-modal strategies of different operators. The Gaussian RL policy is similarly inconsistent and, more critically, exhibits unstable, high-frequency jitter towards the end of the task. In stark contrast, our diffusion RL policy shows extremely low variance, demonstrating that it has converged to a single, optimal strategy. This consistency provides a clear and unambiguous learning signal for the downstream VLA model, which is superior to the noisy signals from human and conventional RL data.

\section{Conclusion}
\label{sec:conclusion}
In this paper, we presented a data generation framework for generating high-quality synthetic data for VLA training using a stabilized diffusion-based RL agent. We demonstrated that the resulting low-variance, optimal trajectories serve as a superior training signal to inconsistent human demonstrations, significantly improving both in-distribution performance and out-of-distribution generalization.

{
	\bibliographystyle{IEEEtran}
	\bibliography{main}

\begin{thebibliography}{10}
\providecommand{\url}[1]{#1}
\csname url@samestyle\endcsname
\providecommand{\newblock}{\relax}
\providecommand{\bibinfo}[2]{#2}
\providecommand{\BIBentrySTDinterwordspacing}{\spaceskip=0pt\relax}
\providecommand{\BIBentryALTinterwordstretchfactor}{4}
\providecommand{\BIBentryALTinterwordspacing}{\spaceskip=\fontdimen2\font plus
\BIBentryALTinterwordstretchfactor\fontdimen3\font minus \fontdimen4\font\relax}
\providecommand{\BIBforeignlanguage}[2]{{%
\expandafter\ifx\csname l@#1\endcsname\relax
\typeout{** WARNING: IEEEtran.bst: No hyphenation pattern has been}%
\typeout{** loaded for the language `#1'. Using the pattern for}%
\typeout{** the default language instead.}%
\else
\language=\csname l@#1\endcsname
\fi
#2}}
\providecommand{\BIBdecl}{\relax}
\BIBdecl

\bibitem{ma2024survey}
Y.~Ma, Z.~Song, Y.~Zhuang, J.~Hao, and I.~King, ``A survey on vision-language-action models for embodied {AI},'' \emph{CoRR}, vol. abs/2405.14093, 2024.

\bibitem{sapkota2025vision}
R.~Sapkota, Y.~Cao, K.~I. Roumeliotis, and M.~Karkee, ``Vision-language-action models: Concepts, progress, applications and challenges,'' \emph{arXiv preprint arXiv:2505.04769}, 2025.

\bibitem{zhong2025survey}
Y.~Zhong, F.~Bai, S.~Cai, X.~Huang, Z.~Chen, X.~Zhang, Y.~Wang, S.~Guo, T.~Guan, K.~N. Lui \emph{et~al.}, ``A survey on vision-language-action models: An action tokenization perspective,'' \emph{arXiv preprint arXiv:2507.01925}, 2025.

\bibitem{oxe}
Q.~Vuong, S.~Levine, H.~R. Walke, K.~Pertsch, A.~Singh, R.~Doshi, C.~Xu, J.~Luo, L.~Tan, D.~Shah \emph{et~al.}, ``Open x-embodiment: Robotic learning datasets and rt-x models,'' in \emph{Towards Generalist Robots: Learning Paradigms for Scalable Skill Acquisition@ CoRL2023}, 2023.

\bibitem{sancaktar2022curious}
C.~Sancaktar, S.~Blaes, and G.~Martius, ``Curious exploration via structured world models yields zero-shot object manipulation,'' \emph{Advances in Neural Information Processing Systems}, vol.~35, pp. 24\,170--24\,183, 2022.

\bibitem{libero}
B.~Liu, Y.~Zhu, C.~Gao, Y.~Feng, Q.~Liu, Y.~Zhu, and P.~Stone, ``Libero: Benchmarking knowledge transfer for lifelong robot learning,'' \emph{Advances in Neural Information Processing Systems}, vol.~36, pp. 44\,776--44\,791, 2023.

\bibitem{zhou2025spire}
Z.~Zhou, A.~Garg, D.~Fox, C.~R. Garrett, and A.~Mandlekar, ``Spire: Synergistic planning, imitation, and reinforcement learning for long-horizon manipulation,'' in \emph{Conference on Robot Learning}.\hskip 1em plus 0.5em minus 0.4em\relax PMLR, 2025, pp. 2347--2371.

\bibitem{mujoco}
E.~Todorov, T.~Erez, and Y.~Tassa, ``Mujoco: A physics engine for model-based control,'' in \emph{2012 IEEE/RSJ international conference on intelligent robots and systems}.\hskip 1em plus 0.5em minus 0.4em\relax IEEE, 2012, pp. 5026--5033.

\bibitem{ye2023reinforcement}
W.~Ye, Y.~Zhang, H.~Weng, X.~Gu, S.~Wang, T.~Zhang, M.~Wang, P.~Abbeel, and Y.~Gao, ``Reinforcement learning with foundation priors: Let embodied agent efficiently learn on its own,'' in \emph{8th Annual Conference on Robot Learning}, 2024.

\bibitem{julg2025refined}
T.~Jülg, W.~Burgard, and F.~Walter, ``Refined policy distillation: From vla generalists to rl experts,'' \emph{CoRR}, vol. abs/2503.05833, March 2025.

\bibitem{luo2025precise}
J.~Luo, C.~Xu, J.~Wu, and S.~Levine, ``Precise and dexterous robotic manipulation via human-in-the-loop reinforcement learning,'' \emph{Science Robotics}, vol.~10, no. 105, p. eads5033, 2025.

\bibitem{shah2022offline}
D.~Shah, A.~Bhorkar, H.~Leen, I.~Kostrikov, N.~Rhinehart, and S.~Levine, ``Offline reinforcement learning for visual navigation,'' in \emph{6th Annual Conference on Robot Learning}, 2022.

\bibitem{tan2025interactive}
S.~Tan, K.~Dou, Y.~Zhao, and P.~Kraehenbuehl, ``Interactive post-training for vision-language-action models,'' in \emph{Workshop on Foundation Models Meet Embodied Agents at CVPR 2025}, 2025.

\bibitem{zhao2025unlocking}
H.~Zhao, W.~Song, D.~Wang, X.~Tong, P.~Ding, X.~Cheng, and Z.~Ge, ``More: Unlocking scalability in reinforcement learning for quadruped vision-language-action models,'' \emph{CoRR}, 2025.

\bibitem{guo2025improving}
Y.~Guo, J.~Zhang, X.~Chen, X.~Ji, Y.-J. Wang, Y.~Hu, and J.~Chen, ``Improving vision-language-action model with online reinforcement learning,'' in \emph{2025 IEEE International Conference on Robotics and Automation (ICRA)}, 2025, pp. 15\,665--15\,672.

\bibitem{zhang2024grape}
Z.~Zhang, K.~Zheng, Z.~Chen, J.~Jang, Y.~Li, S.~Han, C.~Wang, M.~Ding, D.~Fox, and H.~Yao, ``{GRAPE}: Generalizing robot policy via preference alignment,'' in \emph{ICRA 2025 Workshop on Foundation Models and Neuro-Symbolic AI for Robotics}, 2025.

\bibitem{zhai2024fine}
S.~Zhai, H.~Bai, Z.~Lin, J.~Pan, P.~Tong, Y.~Zhou, A.~Suhr, S.~Xie, Y.~LeCun, Y.~Ma \emph{et~al.}, ``Fine-tuning large vision-language models as decision-making agents via reinforcement learning,'' \emph{Advances in Neural Information Processing Systems}, vol.~37, pp. 110\,935--110\,971, 2024.

\bibitem{liu2025can}
J.~Liu, F.~Gao, B.~Wei, X.~Chen, Q.~Liao, Y.~Wu, C.~Yu, and Y.~Wang, ``What can rl bring to vla generalization? an empirical study,'' \emph{arXiv preprint arXiv:2505.19789}, 2025.

\bibitem{xu2024rldg}
C.~Xu, Q.~Li, J.~Luo, and S.~Levine, ``{RLDG:} robotic generalist policy distillation via reinforcement learning,'' \emph{CoRR}, vol. abs/2412.09858, 2024.

\bibitem{o2024open}
A.~O’Neill, A.~Rehman, A.~Maddukuri, A.~Gupta, A.~Padalkar, A.~Lee, A.~Pooley, A.~Gupta, A.~Mandlekar, A.~Jain \emph{et~al.}, ``Open x-embodiment: Robotic learning datasets and rt-x models: Open x-embodiment collaboration 0,'' in \emph{2024 IEEE International Conference on Robotics and Automation (ICRA)}.\hskip 1em plus 0.5em minus 0.4em\relax IEEE, 2024, pp. 6892--6903.

\bibitem{zhou2024genesis}
G.~Authors, ``Genesis: A generative and universal physics engine for robotics and beyond,'' December 2024.

\bibitem{diffusionpolicy}
C.~Chi, Z.~Xu, S.~Feng, E.~Cousineau, Y.~Du, B.~Burchfiel, R.~Tedrake, and S.~Song, ``Diffusion policy: Visuomotor policy learning via action diffusion,'' \emph{The International Journal of Robotics Research}, p. 02783649241273668, 2023.

\bibitem{ho2020denoising}
J.~Ho, A.~Jain, and P.~Abbeel, ``Denoising diffusion probabilistic models,'' \emph{Advances in Neural Information Processing Systems}, vol.~33, pp. 6840--6851, 2020.

\bibitem{diffusionrl}
M.~Makarova, Q.~Liu, and D.~Tsetserukou, ``Diffusionrl: Efficient training of diffusion policies for robotic grasping using rl-adapted large-scale datasets,'' \emph{arXiv preprint arXiv:2505.18876}, 2025.

\bibitem{ren2024diffusion}
A.~Z. Ren, J.~Lidard, L.~L. Ankile, A.~Simeonov, P.~Agrawal, A.~Majumdar, B.~Burchfiel, H.~Dai, and M.~Simchowitz, ``Diffusion policy policy optimization,'' in \emph{CoRL 2024 Workshop on Mastering Robot Manipulation in a World of Abundant Data}, 2024.

\bibitem{schulman2015high}
J.~Schulman, P.~Moritz, S.~Levine, M.~I. Jordan, and P.~Abbeel, ``High-dimensional continuous control using generalized advantage estimation,'' in \emph{International Conference on Learning Representations}, 2016.

\bibitem{schulman2017proximal}
J.~Schulman, F.~Wolski, P.~Dhariwal, A.~Radford, and O.~Klimov, ``Proximal policy optimization algorithms,'' \emph{arXiv preprint arXiv:1707.06347}, 2017.

\bibitem{song2020denoising}
J.~Song, C.~Meng, and S.~Ermon, ``Denoising diffusion implicit models,'' in \emph{International Conference on Learning Representations}, 2021.

\bibitem{loshchilov2017sgdr}
I.~Loshchilov and F.~Hutter, ``Sgdr: Stochastic gradient descent with warm restarts,'' in \emph{International Conference on Learning Representations}, 2017.

\bibitem{pi}
\BIBentryALTinterwordspacing
K.~Black, N.~Brown, D.~Driess, A.~Esmail, M.~Equi, C.~Finn, N.~Fusai, L.~Groom, K.~Hausman, B.~Ichter, S.~Jakubczak, T.~Jones, L.~Ke, S.~Levine, A.~Li-Bell, M.~Mothukuri, S.~Nair, K.~Pertsch, L.~X. Shi, J.~Tanner, Q.~Vuong, A.~Walling, H.~Wang, and U.~Zhilinsky, ``$\pi_0$: A vision-language-action flow model for general robot control,'' 2024. [Online]. Available: \url{https://arxiv.org/abs/2410.24164}
\BIBentrySTDinterwordspacing

\bibitem{openvla}
\BIBentryALTinterwordspacing
M.~J. Kim, K.~Pertsch, S.~Karamcheti, T.~Xiao, A.~Balakrishna, S.~Nair, R.~Rafailov, E.~Foster, G.~Lam, P.~Sanketi, Q.~Vuong, T.~Kollar, B.~Burchfiel, R.~Tedrake, D.~Sadigh, S.~Levine, P.~Liang, and C.~Finn, ``Openvla: An open-source vision-language-action model,'' 2024. [Online]. Available: \url{https://arxiv.org/abs/2406.09246}
\BIBentrySTDinterwordspacing

\bibitem{balasubramanian2015analysis}
S.~Balasubramanian, A.~Melendez-Calderon, A.~Roby-Brami, and E.~Burdet, ``On the analysis of movement smoothness,'' \emph{Journal of neuroengineering and rehabilitation}, vol.~12, no.~1, p. 112, 2015.

\end{thebibliography}
}

\addtolength{\textheight}{-12cm}   

\end{document}